# Dataset of soil images with corresponding particle size distributions for photogranulometry

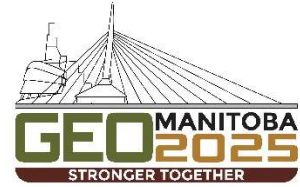

Thomas Plante St-Cyr, François Duhaime, Jean-Sébastien Dubé
*Department of Construction Engineering - École de technologie supérieure (ÉTS), Montreal, Quebec, Canada*
Simon Grenier
*WSP Canada, Montreal, Quebec, Canada*

***Preprint version (Accepted for presentation)***

ABSTRACT
Traditional particle size distribution (PSD) analyses create significant downtime and are expensive in labor and maintenance. These drawbacks could be alleviated using optical grain size analysis integrated into routine geotechnical laboratory workflow. This paper presents a high-resolution dataset of 12,714 images of 321 different soil samples collected in the Montreal, Quebec region, alongside their PSD analysis. It is designed to provide a robust starting point for training convolutional neural networks (CNN) in geotechnical applications. Soil samples were photographed in a standardized top-view position with a resolution of 45 MP and a minimum scale of 39.4 µm per pixel, both in their moist and dry states. A custom test bench employing 13″×9″ white aluminum trays on which the samples are spread in a thin layer was used. For samples exceeding a size limit, a coning and quartering was employed for mass reduction.

RÉSUMÉ
Les analyses granulométriques (PSD) traditionnelles créent des temps morts et sont coûteux en main-d'œuvre et en maintenance. Ces limitations pourraient être atténuées par l'utilisation d'une analyse optique de la granulométrie dans les activités de routine des laboratoires de géotechnique. Cet article présente un ensemble de données de 12 745 images haute résolution de 321 échantillons de sol différents de la région de Montréal, Québec, ainsi de leurs analyses granulométriques. Il est conçu pour fournir un point de départ robuste pour l'entraînement de réseaux de neurones convolutionnels (CNN) dans des applications géotechniques. Il démontre également l'intégration pratique de la granulométrie optique dans les flux de travail des laboratoires géotechniques. Les échantillons de sol ont été photographiés dans une position standardisée en vue de dessus avec une résolution de 45 MP et une échelle minimale de 39,4 µm par pixel, à la fois dans leur état humide et sec. Un banc d'essai sur mesure utilisant des plateaux en aluminium blanc de 13″×9″ sur lesquels les échantillons sont étalés en fine couche a été utilisé. Les échantillons qui dépassent une limite de taille sont divisés par quartage.

## 1 INTRODUCTION

According to the Canadian Foundation Engineering Manual (Canadian Geotechnical Society, 2023), knowing the particle size distribution (PSD) of a soil is crucial for its classification, for predicting its geotechnical properties (e.g. permeability and void ratios), and for selecting appropriate engineering calculations. There are different ways to do a PSD analysis on a granular sample, such as sieving, hydrometry and laser diffractometry. This paper focuses on particles ranging from 80 µm to 80 mm. For this range of particle size, sieving is the most relevant method.

Sieving consists of simply passing a soil sample in a series of sieves with diminishing mesh sizes. The sieves are mechanically shaken for a fixed amount of time (Bureau de normalisation du Québec (BNQ), 2013). The mass retained on each sieve is measured and a cumulative PSD curve is drawn. The cumulative PSD represents the mass percentage of material finer than a given mesh size against the mesh size on a logarithmic scale. The PSD is a monotonically increasing function from 0% to 100%. The increase in cumulative percentage between two consecutive sieve sizes reflects the mass fraction of material retained on the finer of the two sieves.

Traditional PSDs tend to be very precise, and a framework of standards already exists to ensure accurate analysis. However, the main drawback of mechanical sieving is currently the significant amount of time each analysis takes and the manipulations it requires. According to a brief survey of technicians and managers from a commercial geotechnical laboratory, the process of mechanical sieving and its preparatory steps typically require around 2.5 hours of labor distributed over **three days**. The primary delay is the oven drying. It is typically done overnight to ensure the samples are completely dried (BNQ, 2013; ASTM, 2017). Each sample has to be dried twice: a first time to measure the total dry mass and a second time after washing the <5 mm fraction to remove the <80 µm particles.

Historically, image analysis methods for determining PSD fell into two categories. The first involved particle segmentation, while the second analyzed overall image texture. Modern deep learning techniques, such as Convolutional Neural Networks (CNNs), are highly versatile and capable of performing both segmentation and texture analysis. Essentially, CNNs identify patterns from a labeled subset of soil images and extrapolate these learned relationships to predict attributes in any new soil image. The dimensionality of image data is very high, because of the many pixels and soil types. This results in a high number of potential patterns, many of which are not relevant for accurate analysis. It is fundamental to utilize a large volume of images to find the optimal patterns. Otherwise, the CNN will overfit its training data and generalize poorly to unseen images. (Manashti, 2022)

To address the need for extensive data in training advanced machine learning models like CNNs for soil analysis, this study provides the scientific community with a large database of high-resolution soil photographs and their PSDs. The paper first reviews existing image-based methods and datasets. It then describes the custom test bench built for the project and finally presents an overview of the dataset. To our best knowledge, this paper presents the most extensive open database of soil photographs with matching PSD.

## 2 PHOTOGRANULOMETRY: BACKGROUND AND RELATED WORK

### 2.1 Image-based Approach to PSD Analysis

An optical approach to a PSD analysis involves capturing an image with various possible devices like a digital camera and analyzing it with specialized software such as a CNN. This method offers many advantages including being labor and cost effective compared to traditional sieving methods and can be easily implemented for in-situ application in the field.

### 2.2 Motivation for an Image-based Approach Using Machine Learning

Due to time and resource constraints, the current industrial practice for site investigation generally does not include traditional PSD analysis on each collected sample. Sample selection for analysis often includes an element of randomness. In Quebec, the typical procedure involves the engineer going to the lab and selecting the samples he deems critical, but this process requires experience which is not always available. In contrast, having an optical approach to PSD analysis would enable a more informed selection of samples. Unlike purely laboratory-based methods, an optical method can be performed in-situ using minimal equipment. This lowers the cost since it trades the labor of the engineer for the labor of the technician on-site.

Another advantage offered by optical methods is the possibility of having access to high-quality photos of the materials. Unlike physical samples that must be transported and handled, these images can be viewed remotely facilitating collaboration and expert consultation.

A potential limitation of the optical PSD analysis method is lower precision for individual samples. However, because capturing images is both cost and time-effective, it becomes feasible to perform a much greater number of PSD analyses per investigation site. This could allow for the collection of many more soil samples per project, which would provide a more complete soil stratigraphy with a higher resolution.

### 2.3 Related Work: Existing Datasets and Methods

Looking at current datasets for image-based PSD analysis shows they have some important limitations. Comparing datasets can be challenging because some authors treat image tiles as equivalent to whole images, which may not be accurate for representativeness. The reason is that achieving representativeness requires analyzing a sufficiently large sample surface area, not merely the small fractions captured via tiling.

Many previous datasets include a relatively low image count. Among those with more images, McFall et al. (2024) utilized 2458 images tiled from 158 initial images for training a SediNet model, and Gelfenbaum et al. (2017) acquired 2522 images of the Columbia riverbed. However, other studies rely on much smaller datasets. For example, Soranzo et al. (2025) used 136 original images from 26 different samples, which were subsequently tiled to increase the number of training examples.

The methods used to determine the ground truth PSDs vary between datasets. For instance, the PSDs reported by Lang et al. (2021) or Buscombe (2020) were derived from digital image segmentation without corresponding data from physical mechanical sieving of the granular material. Therefore, the PSD predictions of the model do not necessarily match the results obtained with standard tests like sieving. More recent datasets, such as those presented by Duhaime et al. (2021), McFall et al. (2024), and Soranzo et al. (2025), generally incorporate ground truth PSDs obtained through traditional mechanical sieving methods.

Existing datasets often exhibit bias towards specific material types and frequently contain a limited number of distinct samples. For example, the images in Lang et al. (2021) consist mainly of alluvial sediments, Duhaime et al. (2021) presented only 15 different soil samples, which were themselves derived from only 5 initial base samples, and McFall et al. (2023) has a dataset of only beach sands.

The use of very high-resolution images in a dataset enables a wider range of machine learning techniques to be employed. Limited image resolution restrains the PSD analysis to a small range of particle. Before the present work, the Duhaime et al. (2021) dataset, with its 12.2 MP image resolution, represented the upper range available for this type of data. In contrast, the dataset of seafloor sediments images acquired by Gelfenbaum et al. (2017) featured a much lower resolution of 0.37 MP.

### 2.4 CNN Fundamentals and Data Requirements for PSD Prediction

A simple CNN that analyzes an image is made of blocks containing a convolution layer that detects local patterns with filters, an activation function that introduces non-linearity, and often a pooling layer to reduce data size by summarizing features. Additionally, other layers such as a normalization layer can also be added. By stacking these blocks, the model can understand more complex patterns in the data. Following the CNN layers, a Multi-Layer Perceptron (MLP) is often added to perform classification (ex: USCS symbol S or G) or a regression (ex: clay = 16%). Generalizing effectively to unseen examples with this structure requires a large amount of data. For instance,

Helber et al. (2019) demonstrated that having more images allowed models to improve significantly their performance.

The need for large datasets is particularly relevant for PSD prediction, which involves estimating the percentage of material passing each sieve. Ideally, in a balanced dataset, every possible percentage value measured at each sieve should be represented by at least one PSD curve, ensuring that all potential scenarios are covered. In practice, the full range of USCS symbols or PSD curves are not uniformly represented among real-world soils. Instead, real soils typically exhibit patterns where data concentrates within certain regions of the PSD, with outliers appearing sporadically. We calculated that Silty Sands (SM) constituted 78% of sample analyses conducted in one commercial laboratory in Quebec.

Another important consideration is the geometry of the PSD curves. For example, some PSD distributions can be uniforms, while others can be relatively flat or have several plateaus (Chapuis, 2021). Including one example for each possible geometry is theoretically desirable but quantifying this requirement is challenging. For instance, uniform PSDs are rare in natural deposits but common in construction materials.

Alternatively, data engineering can simplify the learning task by changing the prediction target, which may reduce data needs. Models could predict the parameters of a PSD curve equation, instead of directly predicting the percentage passing for each sieve. For example, Soranzo et al. (2025) used a two-parameters Weibull equation for this purpose, though the limited complexity might restrict its applications for curves with multiple plateaus.

## 3 MATERIALS AND METHODS

### 3.1 Experimental Setup

A simple test bench was designed to capture images of 45MP that exhibit minimal variance for distance, angle, lighting, tray color, and camera settings. As shown in Figure 1, it is built from a frame of aluminum extrusions on which a Canon EOS R5 camera with a RF100mm macro lens is fixed on the top portion. A trade-off was necessary between the total sample surface area captured in each photograph and the µm/pixel scale that dictates the minimum distinguishable particle size. Readily available 13x9-inch trays were selected based on trial and error. Because the fixed focal length of the lens dictates scale from tray size, viewing the entire tray for representativeness yielded a 39.4 µm/pixel scale. However, true image scale varies with sample thickness.

To prevent potential glare from the aluminum surface and enhance contrast with typically darker soil samples, the trays were coated with white BBQ paint. This type of paint ensures that the trays can be put in the oven without emitting dangerous fumes. The trays require periodic recoating whenever scratches or clay stains compromise the painted surface. At the bottom of the test bench, a square guide frame facilitates standardized placement of the inserted trays.

Two Amaran COB 60x S LED Bicolor lights of 5500 K and 6000 lumens are installed on both sides of the tray. The light sources are angled to maximize luminosity on the sample and to minimize strong shadows. Another set of smaller lights are positioned in the back and the front.

The camera settings include Fv mode, 1-point AF, 8192×5464 resolution, fixed ISO 100, auto shutter, and fixed f/10 aperture. White balance was calibrated with 96-brilliance white paper under these lights. The camera is connected to a laptop with the "Canon EOS Utility" application for easier capture and management of the images. A wired remote control for the camera helps keep the laptop clean from fine particles between sample mixing.

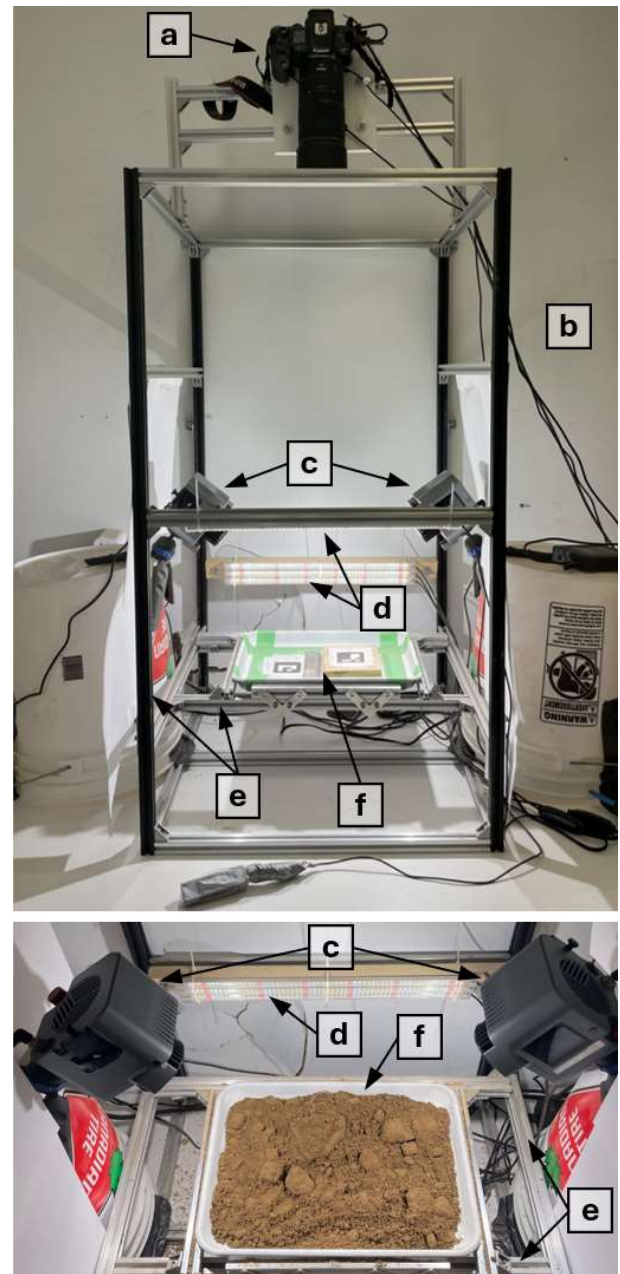

Figure 1. Two different angles of the test bench. a) Canon EOS R5 camera, b) Laptop (out of frame), c) Amaran COB 60x lights, d) Light strips, e) Aluminum frame, f) Removable aluminum tray (top image has calibration tray while bottom image has a soil sample).

### 3.2 Image Calibration

Calibration images were captured using the simple tray depicted in Figure 2 to establish the pixel scale for the samples. Often, a single calibration image served as the reference for multiple samples processed consecutively on the same day. Missing calibration images were replaced with the closest calibration image in terms of time. Variations in sample thickness change the camera-sample distance and impact the µm/pixel scale. On Figure 2, the 50 x 50 mm ArUco markers (black squares) at the bottom of the tray on the right and on a 3/4" plywood on the left have respective scales of 39.4 µm/pixel and 40.2 µm/pixel.

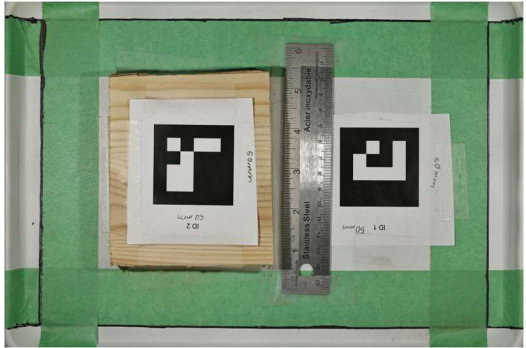

Figure 2. Crude calibration tray

### 3.3 Sample Acquisition

Dry granular materials such as dry soils tend to segregate easily due to the "Brazil nut effect", where larger particles rise to the surface while smaller ones settle beneath them (Rosato et al., 1987). In contrast, from our observations, humid granular materials seem to exhibit significantly less segregation. By reducing particle stacking, thinner samples are hypothesized to minimize segregation effects including gravity-driven sorting and the masking of smaller particles. Consequently, the overall image texture should represent more accurately the actual PSD of the soil.

For this reason, initially, samples weighing between 3 and 5 kg were collected and divided into four trays by using the industry-standard quartering method, following the BNQ 2501-025/2013 standard. The resulting subsamples weighed approximately 1 kg, which was ideal for creating thin layers in the aluminum trays. Classified as "Bulk Samples" in the dataset, most of these samples exhibited a PSD typical of sand with varying proportions of gravel and fine particles, as depicted by the curves presented in Figure 3.

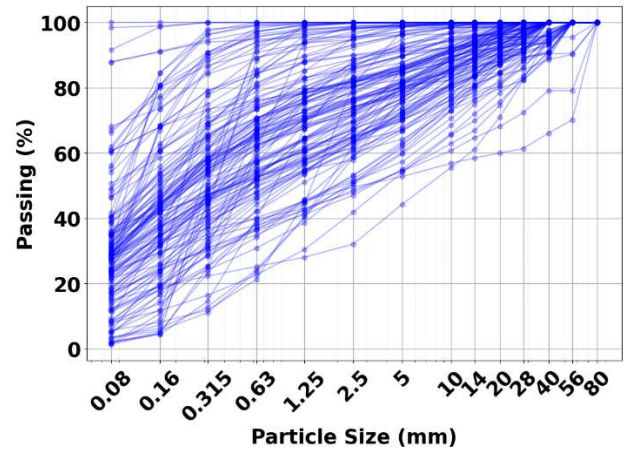

Figure 3. Graph of all Bulk Samples PSD curves

To broaden the types of PSD in the dataset, large samples of uniform gravels from a nearby quarry were collected. By mixing these gravels in various proportions, artificial samples weighing between 2.8 kg and 3.9 kg were created targeting coarser PSD distributions. These samples were also divided with quartering in four trays each. They are classified as "Artificial Samples", with corresponding PSD curves in Figure 4.

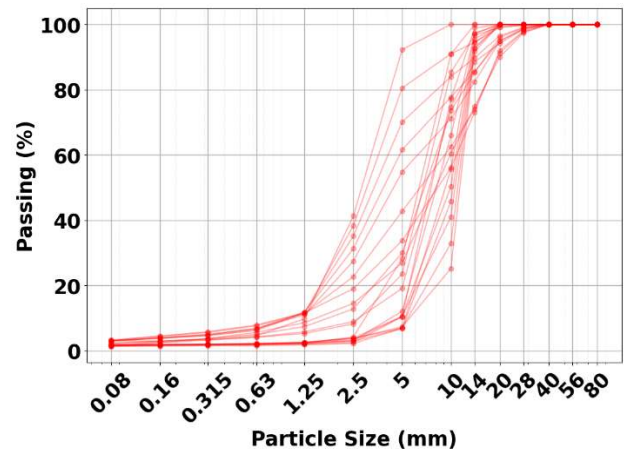

Figure 4. Graph of all Artificial Samples PSD curves

The dataset was later expanded to include samples weighing less than 2 kg that had been collected via the split spoon technique. These samples have PSDs similar to the Bulk samples and are composed primarily of sand with varying proportions of gravel and fine particles. Due to their small mass, they were intentionally kept whole and spread in a single tray rather than being divided. This has the benefit of ensuring that the photograph is representative of the complete sample. They are designated as "Split Spoon Samples", whose PSDs are plotted in Figure 5.

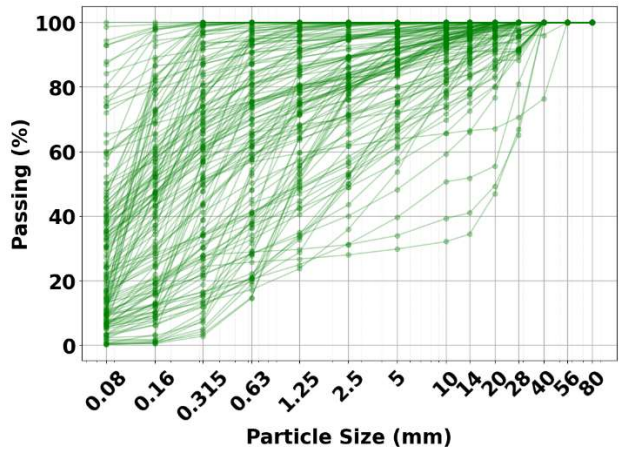
Figure 5. Graph of all split spoon samples PSD curves

According to Chawla et al. (2002), a balanced dataset that includes diverse examples promotes better model generalization, which helps the model recognize even rare cases or outliers. The heatmap in Figure 6 illustrates the concentration of the PSD curves.

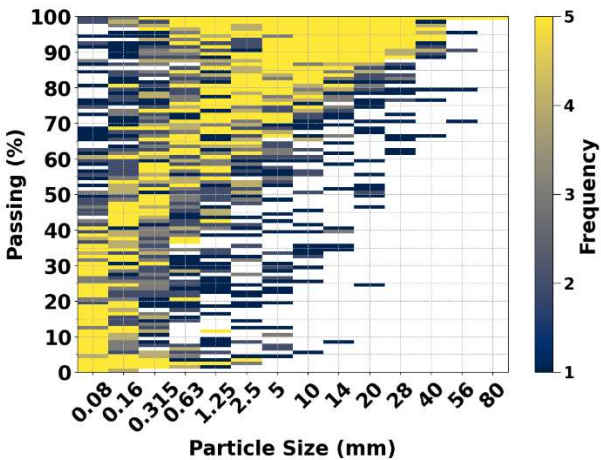
Figure 6. Heatmap with 1% bin interval of all PSD curves of the dataset

### 3.4 Image Acquisition

The data collection process was progressively optimized throughout the study. This section presents the most recent version.

Upon arrival at the laboratory, each sample is prepared individually to minimize moisture loss. The trays used for sample preparation are initially weighed empty. Bulk or artificial sample are divided into four trays using a quartering technique. Split spoon samples are spread into a single tray. The technician then captures five images of each tray in its moist state, mixing the sample between each photograph for data augmentation. In this case, data augmentation is a technique that artificially increases the size of the dataset by physically rearranging the particles. It introduces variation in the images of each sample, which will improve the trained CNN and allow for better generalization. Each moist sample is then weighed in the tray. Next, the trays are placed in an oven and dried overnight. After drying, each tray is photographed five more times, with mixing performed again between each image. Finally, the sample undergoes traditional mechanical sieving in accordance with the BNQ 2501-025/2013 standard.

Some observations were made while developing the data collection process. For sample preparation, a plastic spatula is preferred to spread the soil samples, as it avoids damaging the painted surfaces of the trays. Clays tend to stain the white BBQ paint. The trays can be cleaned with generic melamine foam sponges since they are gentle on the paint. In addition, images should be taken before weighing the soil, especially for clay-rich samples, since the spreading and mixing needed for multiple shots can result in some material loss. It is also important to minimize the time between photographing and weighing moist samples, as they dry quickly either in ambient air or especially under the strong lights of the test bench. Additionally, when humid, soils like silty sands should be photographed as undisturbed as possible, as vibrations can expel water and blur their textural details into a smooth, uniform surface.

### 3.5 Challenges and Lessons Learned

Given the uncertainty about the image resolution needed for accurate geotechnical PSD prediction, our approach focused on maximizing image quality and quantity, while also ensuring a wide variety of soil samples were included.

Initially, pixel shift with the Canon EOS R5 camera was considered, as it can theoretically capture images of 400 MP. However, after multiple attempts, this method was deemed impractical in a geotechnical laboratory environment. A quick and efficient way to capture images was needed, but pixel shift requires a long time between captures and is highly sensitive to vibrations. If interrupted, the process renders the image unusable. Instead, the standard option of 45 MP photographs was used.

At first, every tray was photographed 10 times each under both humid and dry conditions, yielding 80 images per bulk sample. However, it became evident that this was too time-consuming, which limited the overall variety of samples that could be collected. Consequently, the protocol was modified to capture only 5 images per tray. This dropped the number of photos per bulk and artificial samples to 40 and to 10 for split spoon samples, which led to higher sample production

One drawback of dividing samples into four subsamples is the potential loss of representativeness regarding the PSD of the whole sample compared to each subsample. This is why, when split spoon samples of under 2 kg are available, they are prioritized since they do not need division, granting better representativity.

Lights that do not produce heat are recommended to minimize the drying effect on the humid samples.

Clays take longer to process than coarser materials (like sand or gravel) because they spread less easily and form aggregates upon drying that require crushing.

### 3.6 Industry Implementation of Optical PSD

Following model training, the optical PSD method is both fast and simple to operate. Sample preparation and photography may take as little as five minutes, with model predictions running in the background while new samples are processed. Thanks to the straightforward process, technician training is very quick. Additionally, future iterations of the setup could be more compact for easier field applications.

### 4 DATA DESCRIPTION AND SPECIFICATIONS

#### 4.1 Overview

The dataset contains 12,714 images of 321 distinct samples captured in both their naturally moist and oven-dried states and collected in the Montreal region, Quebec, Canada. Each image has a resolution of 8192 × 5464 pixels (44.76 MP), resulting in a total dataset size of more than 425 GB. A traditional mechanical sieving test was performed on all samples to yield PSD data ranging from 80 µm to 80 mm across 14 different sieves, which is provided in an accompanying Excel file. Dry or humid images could not be acquired for some samples due to time constraints within the laboratory. Figures 7, 8, and 9 are image examples for the three sample types.

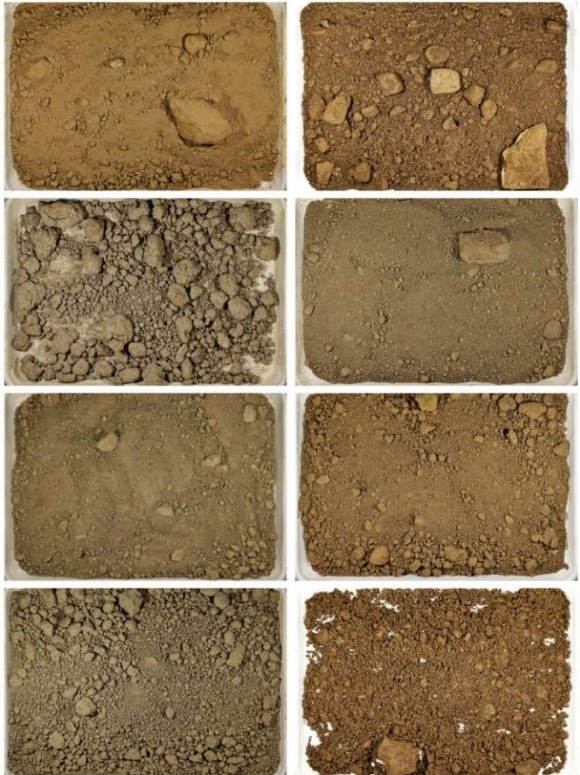

Figure 7. Example of bulk sample images

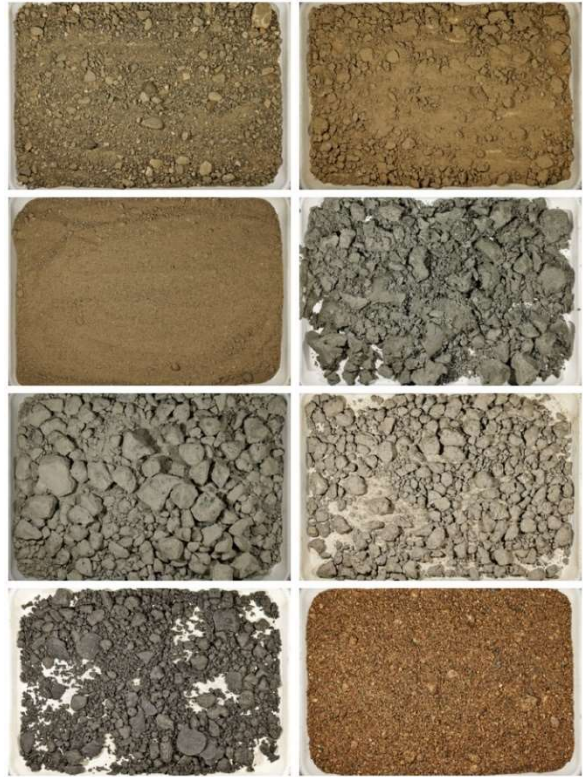

Figure 8. Example of split spoon sample images

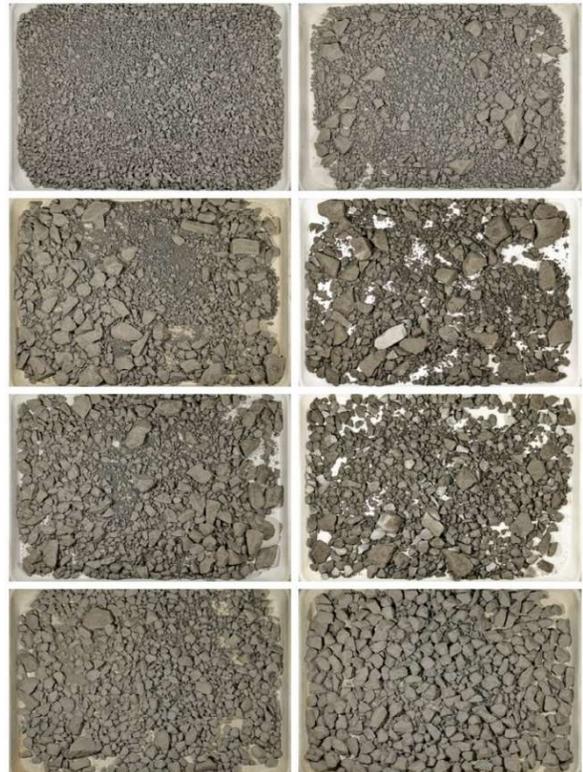

Figure 9. Example of artificial sample images

## 4.2 Dataset comparison

Figure 10 presents a comparative analysis of several relevant datasets in the field using the key metrics of total image count, the number of mechanical sieved sample PSDs, and image resolution. Comparing datasets is difficult because authors sometimes consider image tiles equivalent to whole images, as seen in Lang et al. (2021) who tiled 25 orthophotos of gravel bars. With an initial dataset of only 158 training images, McFall et al. (2024) also employed photo cropping that included overlapping images, which reduces the data's value. On the other hand, Soranzo et al. (2025) writes in his paper that some of his images are 6936 x 9248 pixels, however, the images accessible online have a maximum resolution of 1599 x 1200 pixels. The dataset presented in this paper, highlighted in green, significantly exceeds all other available datasets across every metric, notably without employing cropping or tiling.

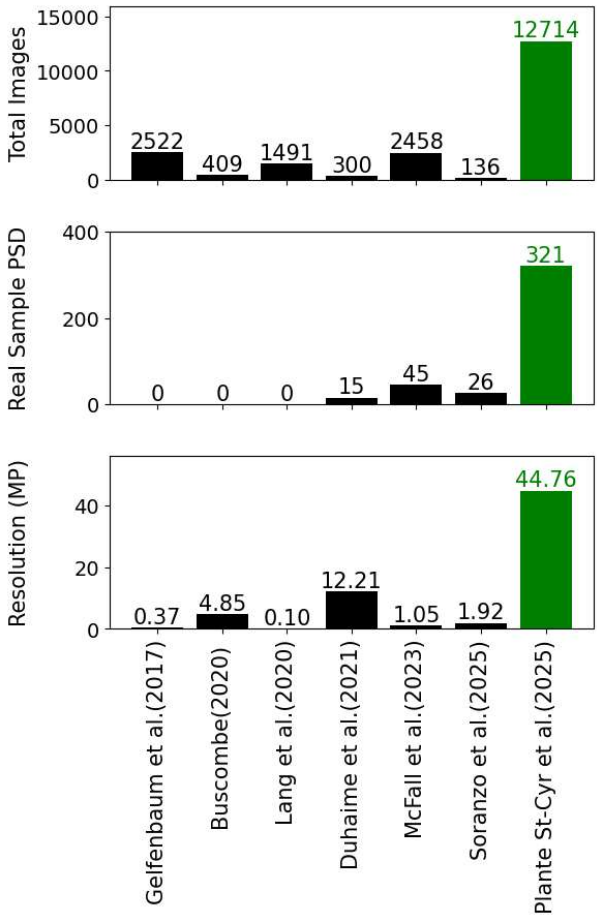

Figure 10. Comparative overview of datasets. Data from this study is highlighted in green.

## 5 DISCUSSION

The significant labor requirements and logistical complexities make it unrealistic to recreate this type of dataset in an academic setting without collaborating with an industry partner. Collaboration could involve requesting access to soil samples that would otherwise be discarded after their analysis are completed. A significant limitation of this approach, however, is that the samples typically lack their original moisture content. While water can be reintroduced, silty and clayey soils necessitate an absorption period, and it remains uncertain whether their characteristics would match those existing before the drying process occurred. A second collaborative approach involves directly embedding research activities within the laboratory's existing workflow, as was done in this research. Successful execution of this model requires meticulous planning in the initial integration phase to mitigate disruption of the laboratory activities. Positioning the photographic stage as the initial step in the operations provides the crucial benefit of capturing the soil samples with their original, undisturbed water content.

## 6 CONCLUSION

This paper introduces a substantial dataset featuring 12,714 high-resolution (45MP) images of 321 soil samples, each paired with 14-point particle size distribution (PSD) ranging from 80 µm and 80 mm obtained from traditional mechanical sieving. Captured under standardized conditions, the dataset serves as an ideal starting point for training machine learning models such as CNNs, for geotechnical particle size distribution analysis. This comprehensive dataset is provided to help drive the transition towards faster, more cost-efficient soil characterization techniques suitable for both standard lab operations and on-site applications.

## 7 DATA AVAILABILITY

The dataset is structured in a main directory 'photogranulometry' housing an Excel file and two image subfolders. The PSD analysis Excel file contains the sample number and the percentage passing through 14 BNQ standard sieve openings (ranging from 80 µm to 80 mm). A 'soil_image' subfolder includes pictures of each sample, where each image filename possesses the sample number, sample type, tray count (1 for split spoon samples, 4 for bulk/artificial), image number (many per tray) and moisture state (humid or dry). A second 'calibration_image' subfolder provides one corresponding calibration image for each soil sample.

The dataset is publicly available on the Federated Research Data Repository (FRDR) of Canada, under the title "Photogranulometry - Dataset of soil images with corresponding particle size distributions". It can be accessed via the following DOI: https://doi.org/10.20383/103.01316.

## 8 ACKNOWLEDGEMENTS

Special thanks to WSP Canada for granting access to their facilities, conducting PSD analyses, and providing technician support for photo capture. We also acknowledge the support of the Natural Sciences and Engineering Research Council of Canada (NSERC) and the Mitacs Accelerate program.